%% file: paper.tex
\pdfoutput=1
\documentclass[11pt]{article}

\usepackage[]{acl}

\usepackage{times}
\usepackage{latexsym}

\usepackage[T1]{fontenc}
\usepackage[utf8]{inputenc}

\usepackage{microtype}

\usepackage{hhline}
\usepackage{booktabs}
\usepackage{multirow}
\usepackage{subfigure}
\usepackage{amsmath}
\usepackage{graphicx}
\usepackage{makecell}
\newcommand{\figref}[1]{Figure \ref{#1}}

\newcommand{\tabref}[1]{Table \ref{#1}}

\newcommand{\apxref}[1]{Appendix \ref{#1}}

\usepackage{color}

\title{Post-Training Dialogue Summarization using Pseudo-Paraphrasing}

\author{Qi Jia$^1$, Yizhu Liu$^1$, Haifeng Tang$^2$, Kenny Q. Zhu$^3$\thanks{\hspace{2mm}The corresponding author.}\\
	$^1$Shanghai Jiao Tong University, Shanghai, China \\
	$^2$China Merchants Bank Credit Card Center, Shanghai, China \\
	\texttt{$^1$\{Jia\_qi, liuyizhu\}@sjtu.edu.cn} \\
	\texttt{$^2$thfeng@cmbchina.com}\\
	\texttt{$^3$kzhu@cs.sjtu.edu.cn}\\
}

\begin{document}
\maketitle
\begin{abstract}
Previous dialogue summarization techniques adapt large language models pretrained on the narrative text by injecting dialogue-specific features into the models. These features either require additional knowledge to recognize or make the resulting models harder to tune. 
To bridge the format gap between dialogues and narrative summaries in 
dialogue summarization tasks, we propose to post-train pretrained 
language models (PLMs) to rephrase from dialogue to narratives. 
After that, the model is fine-tuned for dialogue summarization as usual. 
Comprehensive experiments show that our approach significantly 
improves vanilla PLMs on dialogue summarization and 
outperforms other SOTA models by the summary quality and 
implementation costs.\footnote{Our code and results are publicly available at \url{https://github.com/JiaQiSJTU/DialSent-PGG}.} 

\end{abstract}

\input{intro}
\input{approach}

\input{experiments}

\input{conclusion}

\section*{Acknowledgement}
This research is partially supported by NSFC Grant No. 91646205, and SJTU-CMBCC Joint Research Scheme.

\bibliographystyle{acl_natbib}
\bibliography{anthology,custom}

\clearpage
\newpage
\appendix
\input{appendix}

\end{document}

%% file: intro.tex
\section{Introduction}

\begin{figure}[th]
	\centering
	\includegraphics[width=0.75\columnwidth]{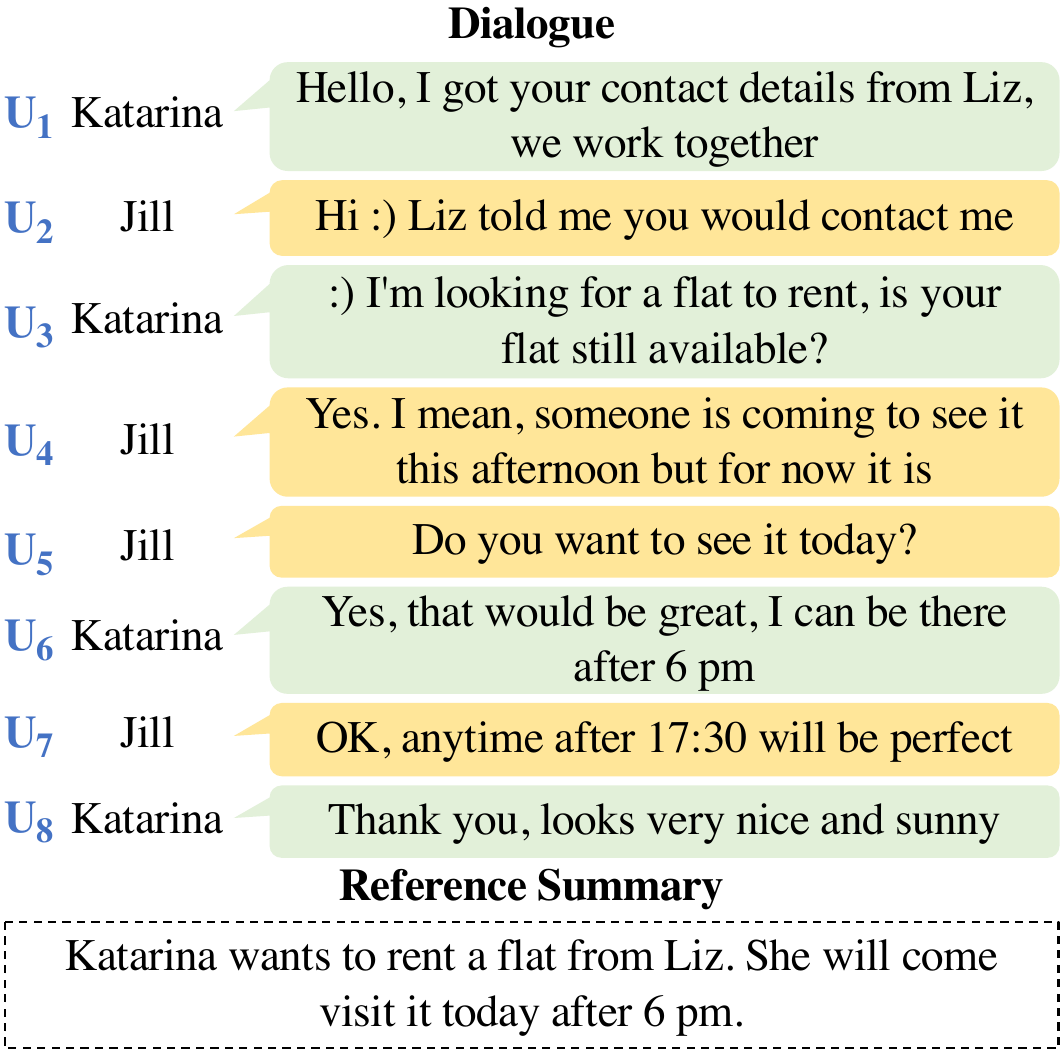}
	\caption{An example from SAMSum dataset. %Different colors are used to identify different speakers.
	}
	\label{fig:example}
\end{figure}

% dialogue summarization task
Dialogue summarization is a specialized summarization task that takes
a series of utterances from multiple speakers in the first person as input,
and outputs fluent and concise summaries in third persons as shown in 
\figref{fig:example}. 
Different from previous monologue inputs such as news~\cite{narayan2018don} and scientific publications~\cite{cohan2018discourse}, dialogues are always less well-organized. They usually contain complicated reference relations, inconsecutive inter-utterance dependencies, informal expressions, and so on, making dialogue summarization a more challenging task.

% pretrain models
The most obvious characteristic of this task is the difference in the format and language styles 
between dialogue and its narrative summary.
Liu, Shi and Chen~\shortcite{liu2021coreference} mentioned
that coreference resolution models trained on general narrative text 
underperforms by about 10\% on dialogue corpus, 
demonstrating the inherent gap between dialogue and narrative text. 
As a result, popular PLMs such as BART~\cite{lewis2020bart} and 
PEGASUS~\cite{zhang2020pegasus} which excel on news summarization 
perform mediocrely on dialogue summarization.

% previous work
To narrow this gap, previous work on dialogue summarization mainly 
resort to injecting dialogue features into PLMs to enhance dialogue 
understanding. These features include dialogue acts~\cite{goo2018abstractive}, topic transitions~\cite{chen2020multi}, coreference relations~\cite{liu2021coreference}, discourse graphs~\cite{chen2021structure}, etc, leading to the rule-based
conversion from dialogues to plain text~\cite{ganesh2019restructuring}.
However, they suffer from three weaknesses. 
First, collecting or extracting these features becomes an additional step
in the summarization pipeline, complicating the inference procedure 
at runtime. 
Second, oracle feature labels are hard to collect and 
errors can propagate from wrong labels to poor summaries.
Third, additional layers or more encoders are required to incorporate 
features into PLMs, increasing the GPU memory footprint both during 
training and inference.

% our approach
A more natural way to bridge this gap is to give the model 
more dialogue-narrative pairs to train on. Due to the scarcity of dialogue
summarization data, one approach~\cite{zhu2020hierarchical} is 
to convert other text summarization
pairs into dialogue to summary pairs via some template, but such work
requires additional data~\footnote{More related work is in \apxref{sec:relatedwork}.}.

In this paper, we propose an alternative approach that 
doesn't use any more data than the original dialogue summarization dataset. 
We convert each existing data pair into many ``\textbf{pseudo-paraphrase}'' pairs 
between a dialogue and a narrative sentence. Then we post-train a pre-trained
seq2seq language model using a \textbf{prefix-guided generation} (PGG) task 
on the augmented paraphrase dataset.
After that, the post-trained model is further fine-tuned as usual 
for dialogue summarization. 
To this end, no human efforts on crafting complicated 
rules or hyper-parameter tuning, or additional memory 
costs, as well as additional training data, is required.
In sum, our contributions are:

\begin{itemize}
\item We propose a novel and effective post-training processing to close 
the format and linguistic style gap between dialogues and narrative texts ($\S$~\ref{sec:approach}).
\item PGG with pseudo-paraphrase pairs requires no extra training data or 
labeling tools for features extractions ($\S$~\ref{sec:pggablation}).
\item Extensive experiments show that the proposed approach compares favorably with current SOTA models using less human efforts and computational costs ($\S$~\ref{sec:end2end}).
\end{itemize}

%% file: approach.tex
\section{Approach}
\label{sec:approach}

The training of a dialogue summarization model is divided two stages: 
post-training and fine-tuning. The model can be any seq-to-seq PLMs
and it remains unchanged
except for the parameters which are updated stage by stage.
We will elaborate on the post-training stage in the rest of this section.

\subsection{Pseudo-paraphrase Dataset Construction}
\label{sec:rephrasedata}

We construct rephrasing datasets from the dialogue summarization dataset 
itself.  
The original dialogue summarization dataset (\textbf{DSum}) 
is made up of dialogue-summary ($D$-$S$) pairs. Each dialogue $D$
is a sequence of utterances and can be concatenated into a whole sequence: 
\begin{equation}
	D = \{U_1, U_2, ..., U_T\} = \{x_1, \ldots, x_n\}
\end{equation}
Each turn $U_t$ is in the form of [$r_t$: $u_t$], where $r$ is a speaker and 
$u$ is the actual utterance. 

Our goal is to create more dialogue to narration kind of paraphrasing pairs. The most
intuitive approach is to divide $S$ into sentences, and pair each sentence
to $D$. We call such pairs ``pseudo-paraphrases'' because the output
sentence (which we call $p$) isn't exactly the paraphrase of the whole 
input, but rather part of the input.%, as $p$ is part of the summary of $D$.

However, doing this poses two challenges: 1) $S$ is a coherent piece of 
text, and its sentences may depend on each other, so a single sentence $p$
out of it may not stand by itself; 2) one $D$ will be paired with
several different $p$, and it is hard for the model to distinguish 
the meaning of these pairs. 

\begin{table}[th]
	\scriptsize
	\centering
	\begin{tabular}{lll}
		\toprule[1pt]
		\textbf{Datasets} & \textbf{Input} & \textbf{Output} \\
		\midrule[1pt]
		{DSum} & $U_{1\sim8}$& \makecell[l]{Katarina wants to rent a flat from Liz.\\ She will come visit it today after 6 pm.}\\		
		\midrule[1pt]
		\multirow{2}{1cm}{{DialSent}} & $U_{1\sim8}$ &\textit{\underline{Katarina} wants} to rent a flat from Liz. \\
		\cmidrule{2-3}
		&$U_{1\sim8}$ &\textit{\underline{Katarina} will come} visit it today after 6 pm. \\
		
		\bottomrule[1pt]
	\end{tabular}
	\caption{Example pseudo-paraphrase pairs generated from the example in Figure~\ref{fig:example}. One pair in DSum becomes two pairs in DialSent. The prefix tokens determined by linguistic features, NOUN and ROOT, are underlined and italic respectively.}
\label{tab:datasets}
\end{table}

To solve 1) we apply coreference resolution\footnote{We use \url{https://spacy.io/}.} on $S$ and convert every pronoun in it to the full reference 
first, before splitting the summary $S$ into sentences. 
Sentence with fewer than 3 words (e.g., ``Ally agree'') 
are discarded since it carries too little information.  
The set of data pairs thus created is called (\textbf{DialSent}). 
An example is in 
\tabref{tab:datasets}.

To tackle 2), one obvious thought is to further split $D$ into sets of sentences
in which each set corresponds to a sentence $p$ in the summary.
However, our extensive experiments (see \apxref{sec:para}) showed that none of the 
straight-forward heuristics work well to establish such alignments.
This is mainly due to the fact that dialogue utterances are highly dependent. Thus, splitting operations are not optimal.
Instead of changing $D$, we decide to use the pseudo-paraphrases
directly but introduce a prefix-guided generation task to guide the model
learning to extract relevant information from $D$.

\subsection{Prefix-guided Generation Task}
\label{sec:pgg}

Summarization for dialogues focuses on analyzing ``who-did-what'' storylines~\cite{chen2021structure} and the beginning of each summary sentence are usually different speakers or the same speaker doing different things. As a result,  using the prefix made up of ``who'' or ``who-did'' can help to select the related information from dialogues or plan the content to be generated.

\begin{figure}[th]
	\centering
	\includegraphics[width=0.75\columnwidth]{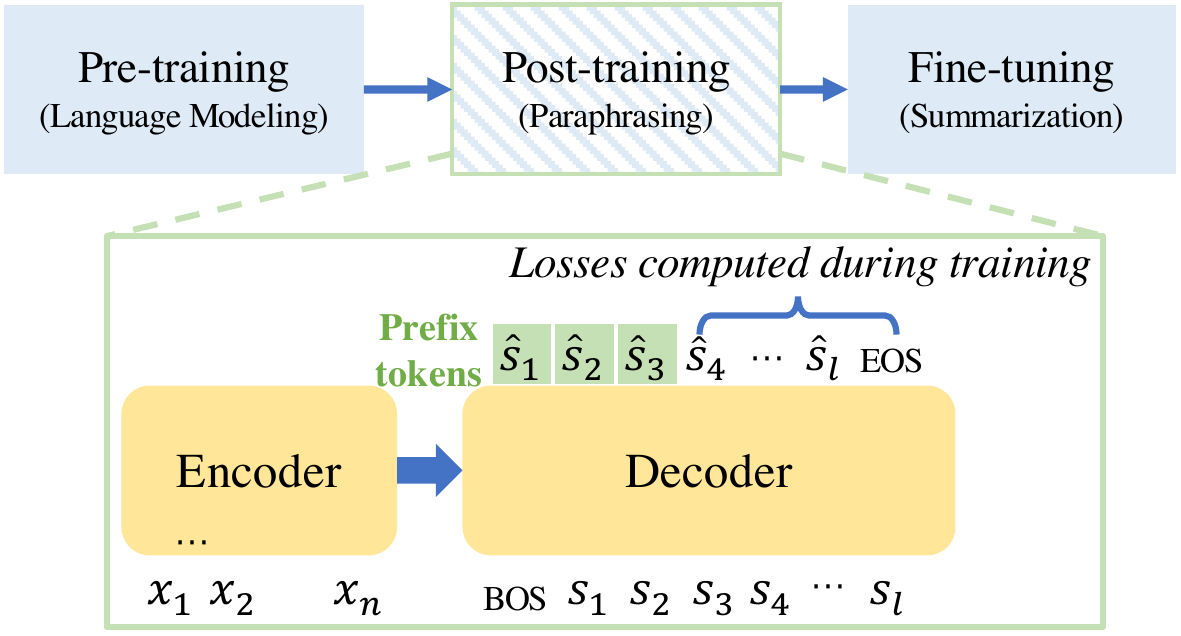}
	\caption{A illustration of our approach. 
BOS and EOS stand for begin and end of the sequence.}
	\label{fig:pgg}
\end{figure}

In other words, we take the inspiration from content 
planning~\cite{narayan2021planning,wu-etal-2021-controllable}. 
When training, the first few tokens of $p$ are provided as prefix to 
the decoder. This prefix serves as an information selection hint to 
the model so it is easier to learn why that particular $p$ should be 
generated. 
The losses are calculated between the generated tokens and reference 
tokens after the prefix as shown in Figure~\ref{fig:pgg}. 

Let $p=\{s_1, \ldots, s_l\}$. 
Our prefix-guided training task is a vanilla auto-regressive generation
task minimizing the negative log-likelihood of $p$:
\begin{equation}
	\begin{aligned}
		L &= -\frac{1}{l-a}\sum_{t=a}^{l}\log P(s_t|s_{<t},H^d) \\
	\end{aligned}
\end{equation}
where $a$ is the number of prefix tokens. $H^d$ is the output hidden vectors of the encoder with input $D$.

There are various ways to determine the prefix length $a$. We can take
a fixed length, a random length or a prefix up to a certain linguistic feature
such as NOUN, VERB or ROOT. 
The exact linguistic feature to use is a dataset-dependent hyper-parameter and 
can be tuned by the validation set. Examples of prefix tokens is marked in Table \ref{tab:datasets}.

%% file: experiments.tex
\section{Evaluation}
We first present the experimental setups, then conduct an ablation study to
determine the proper prefix in PGG training, before our main 
results. More implementation details are in \apxref{sec:details}.

\subsection{Experimental Setup}
We implement our experiments on \textbf{SAMSum}~\cite{gliwa2019samsum}
and  \textbf{DialSumm}~\cite{chen-etal-2021-dialogsum}, whose statistics 
are listed in Table \ref{tab:sumdataset}.

\begin{table}[th]
	\scriptsize
	\centering
	\begin{tabular}{p{0.8cm}rrrrr}
		\toprule[1pt]
		\textbf{Datasets}& \textbf{Variation} & \textbf{Train/Val/Test} & \textbf{IW} & \textbf{OW} & \textbf{CR} \\
		\midrule[1pt]
		\multirow{2}{*}{SAMSum}& DSum & 14,731/818/819 & 124.10 & 23.44 & 0.25\\
		&DialSent& 29,757/1,654  & 149.93 & 11.93 & 0.13\\
		\multirow{2}{*}{DialSumm} & DSum & 12,460/500/500 & 187.52 & 31.02 & 0.18 \\
		& DialSent&22,407/840  &214.00 & 17.78 & 0.10 \\
		\bottomrule[1pt]
	\end{tabular}
	\caption{Statistics of dialogue summarization datasets. IW, OW and CR represent the number of input words, the number of output words and compression ratio (OW/IW) respectively.}
	\label{tab:sumdataset}
\end{table}

We compare our method with these baselines.
\textbf{Lead-3} and \textbf{Longest-3} are simple rule-based 
baselines that extract the first or the longest $3$ utterances in 
a dialogue as the summary respectively. 
\textbf{PGN}~\cite{see2017get}, \textbf{Fast-Abs}~\cite{chen2018fast}, and \textbf{PEGASUS}~\cite{zhang2020pegasus} are well-known models for text summarization. \textbf{BART}~\cite{lewis2020bart} is a general PLM and performs well after fine-tuning.
\textbf{CODS}~\cite{wu-etal-2021-controllable}, \textbf{Multi-view}~\cite{chen2020multi} and \textbf{DialoBART}~\cite{feng-etal-2021-language} are the SOTA models designed for dialogue summarization.

We evaluate both automatically and by human.
For \textbf{automatic evaluation}, we use Rouge-1, 2, and L~\cite{lin2004rouge} 
F1-scores~\footnote{\url{https://pypi.org/project/py-rouge/}}. 
Following \citet{feng-etal-2021-language}, we adopt the same Rouge evaluation tool and compute between reference summaries and generated summaries. For DialSumm, we use maximum rouge scores among references for each sample.
For \textbf{human evaluation}, we three proficient English speakers to evaluate 100 
random samples from SAMSum.  
Each original dialogue and its reference summary are shown with generated summaries in a random order simultaneously. Showing summaries from different approaches together helps humans do comparisons between them.
Following \citet{chen2020multi} and \citet{liu2021coreference}, 
each summary is scored on the scale of $[2, 0, -2]$, where $2$ means concise 
and informative, $0$ means acceptable with minor errors, and $-2$ means 
unacceptable. The final scores are averaged among annotators.
We also ask human annotators to label the error types in the summary. 
We consider the following 4 error types:
\textbf{Mis}sing important contents, 
\textbf{Red}undant content, 
\textbf{Cor}eference mismatches, and
\textbf{Rea}soning error.
Rea and Cor concentrate on comparisons to the dialogue, 
and the rest two focus on comparisons to the reference. 
We determine the error for each case by majority voting, and count the errors of each model.

\subsection{Ablations Study}\label{sec:pggablation}

We conduct ablations to verify the effectiveness of post-training on DialSent with PGG, including post-training on DSum with PGG task (DSum-PGG), DSum with vanilla generation task (DSum-VG), and DialSent with vanilla generation task (DialSent-VG) in Table~\ref{tab:pggablation}.
The results of DSum-VG drop, indicating that fine-tuning for BART on DSum with early-stop is enough. Post-training with the same data and task leads to overfitting. 
DialSent-PGG performs best for two reasons. Compared with DialSent-VG, the prefix solves one-to-many mappings between a dialogue and summary sentences, so that the same dialogue can lead to different generations.
On the other hand, 
the prefix can manipulate the selection within a short sentence but is not strong enough to direct content in multiple sentences. Thus, DialSent-PGG learns more cross-format paraphrasing ability and performs better.

\begin{table}
	\scriptsize
	\centering
	\begin{tabular}{lccc}
		\toprule[1pt]
		\textbf{Models} & \textbf{Rouge-1} & \textbf{Rouge-2} & \textbf{Rouge-L} \\
		\midrule[1pt]
		\multicolumn{4}{l}{\textit{SAMSum}} \\
		%\textbf{BART}  &52.06 &27.45 &48.89  \\
		{DSum-VG} &51.48 &27.27 &49.45 \\
		{DSum-PGG} &52.52 &27.51 & 49.03\\
		{DialSent-VG} &52.16 &27.79 & 49.41\\
		{DialSent-PGG} &\textbf{53.54} &\textbf{28.91} &\textbf{50.21}\\
		\midrule[1pt]
		\multicolumn{4}{l}{\textit{DialSumm}} \\
		%\textbf{BART} &  53.01&29.18 &51.34  \\
		{DSum-VG} &53.15 &28.86 & 51.48\\
		{DSum-PGG} &53.27 &28.64 &51.69 \\
		{DialSent-VG} &52.99 & 29.14& 51.40\\
		{DialSent-PGG} &\textbf{54.73} & \textbf{30.47}&\textbf{53.46} \\
		\bottomrule[1pt]
	\end{tabular}
	\caption{Ablations on DialSent with PGG task.}
	\label{tab:pggablation}
\end{table}

\begin{table}[th]
	\scriptsize
	\centering
	\begin{tabular}{llll}
		\toprule[1pt]
		\textbf{Models} & \textbf{Rouge-1} & \textbf{Rouge-2} & \textbf{Rouge-L} \\
		\midrule[1pt] 
		\multicolumn{4}{l}{\textit{SAMSum}} \\
		{w/o} & 52.16 & 27.79& 49.41\\
		{const} & 51.71& 27.34&49.25 \\
		{random} & 52.32& 27.99&49.68 \\
		{Ling-Noun} &\textbf{53.54} & \textbf{28.91}&50.21\\
		\midrule[1pt]
		\multicolumn{4}{l}{\textit{DialSumm}} \\
		{w/o} & 52.99& 29.14&51.40 \\
		{const} &53.29 &29.57 &52.10 \\
		{random} &53.82 & 29.88&52.43\\
		{Ling-Root} &\textbf{54.73} & \textbf{30.47}&\textbf{53.46}\\
		\bottomrule[1pt]
	\end{tabular}
	\caption{Ablations on prefix designs for PGG.}
	\label{tab:prefixablation}
\end{table}

We try several choices of prefix length:
(1) \textbf{W/O}: without any prefix.
(2) \textbf{Const}: Constant length set to 2 and 3 for SAMSum and
DialSumm respectively, since a person's name is $1.69\pm0.69$ tokens long
on average~\footnote{DialSumm normalizes speaker names 
into ``\#Person1\#'' resulting in more tokens.}.
(3) \textbf{Random}: set by uniform sampling from a range of numbers. 
We set the range to $1\sim3$ and $2\sim4$ for the two datasets respectively.
(4) \textbf{Ling}: using the validation set, we determined that Noun and Root 
are the best choice for the two datasets, respectively. In this way, the number of prefix tokens for SAMSum and DialSum are $1.90\pm1.10$ and $3.55\pm1.24$.

In \tabref{tab:prefixablation}, Ling performs the best among these variants.
The actual linguistic feature to use may vary from dataset to dataset though.
The remaining experiments will be conducted using PGG-Ling.

\begin{table}[th]
	\scriptsize
	\centering
	\begin{tabular}{lccc}
		\toprule[1pt]
		\textbf{Models} & \textbf{Rouge-1} & \textbf{Rouge-2} & \textbf{Rouge-L} \\
		\midrule[1pt]
		\multicolumn{4}{l}{\textit{SAMSum}} \\
		%\textbf{BART}  &52.06 &27.45 &48.89  \\
		{Lead-3} & 31.41& 8.68&30.38 \\
		{Longest-3} &32.46 &10.27 &29.92 \\
		{PGN} &40.08 &15.28 &36.63 \\
		{Fast-Abs} &41.95 &18.06 &39.23 \\
		{PEGASUS}& 50.50 & 27.23 & 49.32 \\
		{BART$^\dag$} &52.06 &27.45 &48.89 \\
		{CODS} &52.65 &27.84 &50.79 \\
		{Multi-view} & 53.42& 27.98& 49.97\\
		{DialoBART} &\textbf{53.70} &28.79 &\textbf{50.81} \\
		{DialSent-PGG$^\dag$} &\underline{53.54} &\textbf{\underline{28.91}} &\underline{50.21} \\
		\midrule[1pt]
		\multicolumn{4}{l}{\textit{DialSumm}} \\
		%\textbf{BART} &  53.01&29.18 &51.34  \\
		{Lead-3} &31.15 &10.08 &30.68\\
		{Longest-3} &27.00 &9.41 &25.31 \\
		{BART$^\dag$} & 53.01& 29.18& 51.34\\
		{DialoBART$^\dag$} &53.26 &29.58 &52.01 \\
		{DialSent-PGG$^\dag$} &\textbf{\underline{54.73}} &\textbf{\underline{30.47}} &\textbf{\underline{53.46}} \\
		\bottomrule[1pt]
	\end{tabular}
	\caption{Dialogue summarization results compared with baselines. $\dag$ represents the models implemented by ourselves. \underline{Underlined} scores are statistically significantly better than BART with $p<0.05$ based on t-test. }
	\label{tab:end2end}
\end{table}

\subsection{Comparison to SOTA Models}\label{sec:end2end}

\begin{figure}
	\centering
	\includegraphics[scale=0.45]{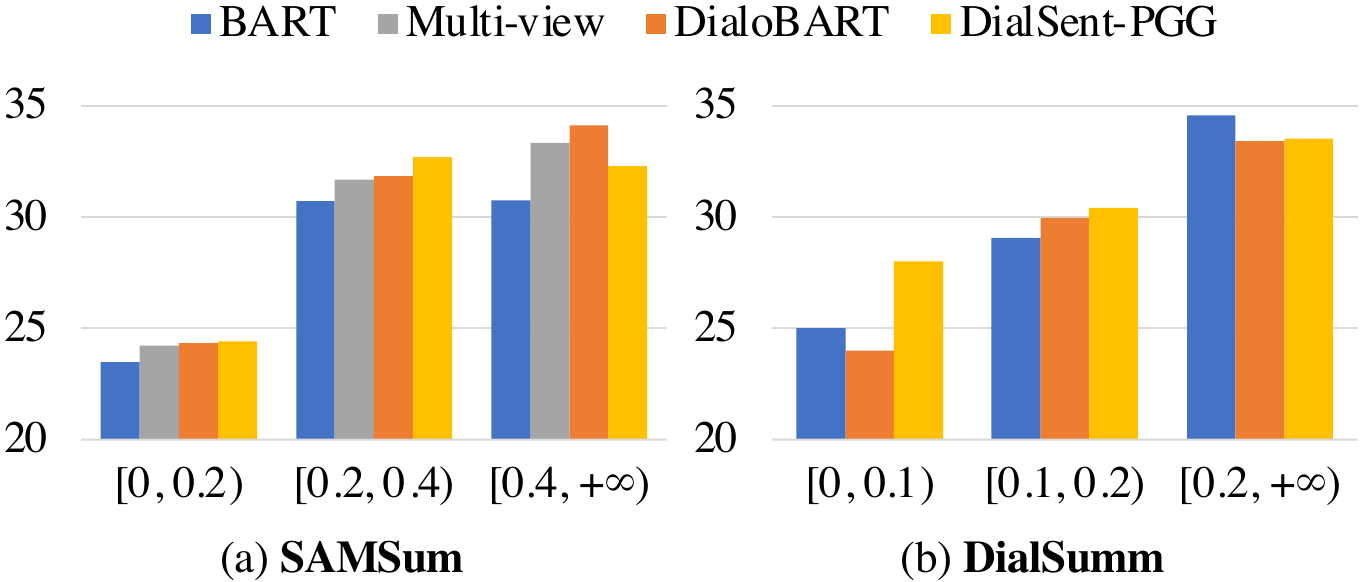}
	\caption{Comparison for models on samples with different CR. X-axis represents the ranges for CR(\%). Y-axis is the Rouge-2 F1(\%).}
	\label{fig:cr}
\end{figure}

\textbf{Automatic Evaluation:} 
Our model DialSent-PGG performs competitively against other models on 
SAMSum and significantly better than the peers on DialSumm. 
It improves 1.5 on Rouge scores over BART for both datasets, 
while DialoBART achieves less gains on DialSumm. 
Based on Table~\ref{tab:datasets}, DialSumm is a more difficult dataset 
with lower compression ratios. Our model performs better on samples 
with lower CR, i.e. more compressed samples, as shown in Figure~\ref{fig:cr}, thus differences 
between DialSent-PGG and DialoBART are more obvious on DialSumm. 
A simple case study is shown in Table~\ref{tab:case}.  
Multi-view faces the repetition problem as it takes the dialogue as 
input twice with two encoders. DialoBART has reasoning errors because 
it regards ``William'' as a keyword. DialSent-PGG instead generates a 
concise and correct summary. More cases are in \apxref{sec:cases}.

\begin{table}[th]
	\scriptsize
	\centering
	\begin{tabular}{lp{4.8cm}}
		\toprule[1pt]
		 {Dialogue}& \makecell[l]{William: are you still angry? \\Emilia: YES  \\William: :(} \\
		 \hline
		 {Multi-view}& Emilia is still angry \textit{and still angry}. \\
		 \hline
		 {DialoBART}& \textit{William and} Emilia are still angry.\\
		 \hline
		 {DialSent-PGG} &Emilia is still angry. \\
		\bottomrule[1pt]
	\end{tabular}
	\caption{A case from SAMSum. \textit{Errors} are in italic.}
	\label{tab:case}
\end{table}

\textbf{Human Evaluation:} The overall human scores on BART, Multi-view, DialoBART and DialSent-PGG are $0.35$, $0.40$, $0.43$ and $0.55$ respectively. 
The Fleiss Kappa among three annotators is $0.39$~\footnote{Fleiss Kappa between 0.4 and 0.6 is considered moderate.}.
The latter three models all improve BART, 
with DialSent-PGG topping the ranks.

\begin{figure}[th]
	\centering
	\includegraphics[scale=0.4]{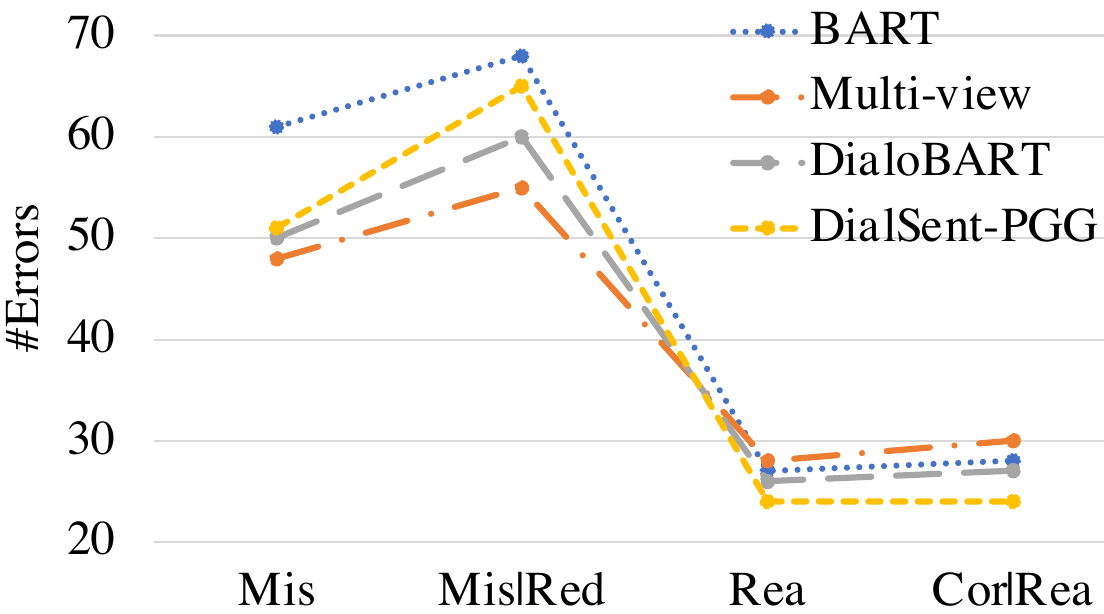}
	\caption{Error analysis on SAMSum. }
	\label{fig:humaneval}
\end{figure}
For error analysis, the Fleiss Kappa for Mis, Red, Cor and Rea are 
$0.55$, $0.10$, $0.26$, $0.42$ respectively. The agreement on Red is 
lower because identifying unimportant information is hard. 
The agreement on Cor is fair due to undistinguishable errors. For example, mismatching of a person and an event among multiple utterances can be either a Cor or a Rea. Besides, Red always leads to Mis. So, 
we divide the error types into two groups and merge them with "OR" logical operation within each group. The Fleiss Kappa for Mis|Red and Cor|Rea are $0.45$ and $0.46$.
We show error types with the agreement larger than $0.40$ in Figure~\ref{fig:humaneval}. 

Multi-view performs better on content selection and 
DialSent-PGG performs better on reasoning and coreference understanding, 
while DialoBART lies in between. 
Fewer errors on Rea and Cor|Rea reflect that our approach 
successfully narrows the understanding gap.
Because references are not the only good summary, high missing content 
doesn't mean that the generated summary is unacceptable. 
As a result, the model with fewer Cor|Rea errors receives higher overall score.

\textbf{Implementation Costs:} 
We compare the implementation costs between our approach and two state-of-the-art models, i.e. Multi-view and DialoBART, in Table~\ref{tab:end2endcost}
Although explicitly injecting features for dialogue understanding is effective, labels for these features are hard to collect and implementation costs for these approaches on a new dataset are high. 
Multi-view and DialoBART proposed doing labeling automatically with unsupervised algorithms or language models. However, these labeling approaches bring extra hyper-parameters which are different between datasets and need to be found by trial and error. If we use the same keywords extraction ratio, similarity threshold and topic segmentation ratio from SAMSum directly, the results on DialSumm are only 50.61/26.67/49.06 (Rouge-1/2/L). We searched for the best combination of hyper-parameters following their paper and did 14 trials, while applying our approach on DialSumm only need 4 trials.

On the other hand, injecting features increases the requirement of GPU memory. 
With the same training parameters(max tokens$=$1024, batch size$=$1, gradient checkpointing$=$False), Multi-view with double-encoder design encounters an out-of-memory error on RTX 2080Ti with 11G GPU memory. DialoBART occupies around 10.36G since it lengthens the dialogue with additional annotations. DialSent-PGG only occupies 9.87G during post-training for recording the length of the prefix, and 9.65G during fine-tuning which is the same as vanilla BART.  
In a word, our approach costs less for implementation.

\begin{table}[th]
	\scriptsize
	\centering
	\begin{tabular}{lcccc}
		\toprule[1pt]
		\textbf{Models} & \textbf{Mem} & \textbf{\#HP}& \textbf{\#Tri}&\textbf{\#St} \\
		\midrule[1pt]
		{Multi-view} &OOM &5 &- &-\\
		{DialoBART} &10.36G &3 &14 &38.61k\\
		{DialSent-PGG} &9.87G/9.65G&1 &4 & 19.32k\\
		\bottomrule[1pt]
	\end{tabular}
	\caption{The upper-bound of GPU memory footprint (Mem), newly introduced hyper-parameter counts (\#HP), the number of trails (\#Tri) and total training steps (\#St) for implementing different models.}
	\label{tab:end2endcost}
\end{table}

%% file: conclusion.tex
\section{Conclusion}

We propose to post-train dialogue summarization models to enhance their 
cross-format rephrase ability by prefix-guided generation training on 
dialogue-sentence pseudo-paraphrases, and get promising results. 
Creating self-supervised tasks for cross-format post-training and 
incorporating compatible features for downstream fine-tuning are 
plausible future directions.

%% file: appendix.tex
\section{Related Work}
\label{sec:relatedwork}
%Consideration of our approach over previous approaches on 
Dialogue summarization and pretrained language models are discussed as follows.

\textbf{Dialogue Summarization:} A growing number of works have been proposed for dialogue summarization in recent years. 
In this work, we mainly refer to the chat summarization defined in~\cite{feng2021survey}.
Previous works widely explore dialogue features explicitly and input them as known labels to enhance the dialogue understanding ability of summarization models.
Features, including dialogue acts~\cite{goo2018abstractive}, topic transitions~\cite{chen2020multi}, discourse dependencies~\cite{chen2021structure}, coreference relations~\cite{liu2021coreference}, argument graphs~\cite{fabbri2021convosumm}, semantic structures or slots~\cite{lei-etal-2021-finer-grain,zhao-etal-2021-give-truth}, etc. are carefully designed and collected by transferring tools pre-trained on other corpus or unsupervised methods with multiple hyper-parameters. 
%Such specific features are not suitable for all scenarios and they highly relied on human labors or experts to redesign heuristic rules or hyper-parameters when implementing on a new dataset.
These work also modify the basic transformer-based models with additional encoders~\cite{chen2020multi} or attention layers~\cite{chen2021structure,liu2021coreference,lei-etal-2021-finer-grain,zhao-etal-2021-give-truth} to utilize the injected features.
\citet{liu-etal-2021-topic-aware} propose a contrastive learning approach for dialogue summarization with multiple training objectives. They also introduce a number of hyper-parameters for contrastive dataset construction and balancing among those objectives.

\textbf{Pretrained Language Models:} Previous pretrained seq-to-seq models can be divided into two categories by training data formats.
One is models pretrained on narrative text, such as BART~\cite{lewis2020bart}, PEGASUS~\cite{zhang2020pegasus}, and T5~\cite{raffel2020exploring}. They use training data from Wikipedia, BookCorpus~\cite{zhu2015aligning} and C4~\cite{raffel2020exploring}. These models show great potentials for tasks such as translation and story ending generation. %, where both the input and output are in narrative text format.
The other is models pretrained on dialogue, such as DialoGPT~\cite{zhang2020dialogpt} and PLATO~\cite{bao2020plato}. Their training data are general-domain dialogues, such as Reddit~\cite{henderson2019repository} and Twitter~\cite{cho2014learning}. These models work for dialogue response selection and generation tasks. %, aiming at finding or generating the most suitable utterance given the dialogue history.
All of the above models are trained to exploit language features within the same data format, with pre-training tasks such as masked token/sentence prediction and utterance permutation.
%on tasks including masked language modeling, next utterance or sentence prediction, and utterance or sentence permutation, exploiting the language modeling features within the same data format.
Pretraining with cross-format data hasn't been researched so far. 
As a first step, we focus on narrowing the gap by learning to rephrase unidirectionally from dialogue to narratives.

\section{Implementation Details}
\label{sec:details}
We use BART\footnote{\url{https://huggingface.co/facebook/bart-large}} as our basic language model. For both post-training and fine-tuning, the speakers and utterances of each dialogue are concatenated into a single sequence and truncated to the first $1024$ tokens.
The learning rate is set to $3e-5$ with weight decay equaling $0.01$. The number of warmup steps is $500$ and dropout is $0.1$. The model is tested on the corresponding validation set after each training epoch and % It stops training when the Rouge-2 F1 score doesn't improve on the validation set or it reach the maximum training epoch.
the early-stop is activated if there is no improvement in the Rouge-2 F1 score.
The early-stop and maximum training epochs are set to $3$ and $10$.
During inference, i.e., validation and testing,
the beam size is set to $4$ with length penalty equaling $1.0$
and no-repeat-n-gram size equaling $3$. 
The minimum and maximum lengths are set to the corresponding lengths
of the reference summaries based on statistics of each dataset, allowing for free-length text generation.
Besides, for the inference on the validation set during the post-training stage, we also set the first $3$ tokens as the known prefix. This constant number enables a fair comparison of performances on validation sets under different experimental settings.
All of our experiments are done on an RTX 2080Ti with 11G GPU
memory. We run experiments three times and show the best results following~\cite{feng-etal-2021-language}. 

\section{Other Types of Paraphrase Datasets}
\label{sec:para}

\begin{table}
	\scriptsize
	\centering
	\begin{tabular}{lll}
		\toprule[1pt]
		\textbf{Datasets} & \textbf{Input} & \textbf{Output} \\
		\midrule[1pt]
		{DialIndirect} & $U_{1\sim8}$& \makecell[l]{Katarina says,``Hello, I got ...\\ we work together'' Jill says,\\``Hi :) ......  nice and sunny''}\\
		\midrule[1pt]
		{ExtSum} &$U_3$, $U_6$ & \makecell[l]{Katarina ...... a flat from Liz.\\ She will ...... after 6 pm.} \\
		\midrule[1pt]
		{ExtSumM} &$U_{3\sim6}$&\makecell[l]{Katarina ...... a flat from Liz.\\ She will ...... after 6 pm.}  \\
		\midrule[1pt]
		\multirow{2}{1cm}{{ExtSent/\\ExtSentM}}&$U_3$&Katarina ...... a flat from Liz. \\
		\cmidrule{2-3}
		& $U_6$ &Katarina will ...... after 6 pm. \\
		\midrule[1pt]
		{DSum} & $U_{1\sim8}$& \makecell[l]{Katarina ...... a flat from Liz.\\ She will ...... after 6 pm.}\\
		
		\midrule[1pt]
		\multirow{2}{1cm}{{DialSent}} & $U_{1\sim8}$ &Katarina ...... a flat from Liz. \\
		\cmidrule{2-3}
		&$U_{1\sim8}$ &Katarina will ...... after 6 pm. \\
		
		\bottomrule[1pt]
	\end{tabular}
	\caption{An illustration of post-training pairs generated from the example in Figure~\ref{fig:example}. ExtSent and ExtSentM get the same training pairs in this case.}
	\label{tab:datasets2}
\end{table}

\begin{table}[h]
	\scriptsize
	\centering
	\begin{tabular}{lrrrr}
		\toprule[1pt]
		\textbf{Datasets} & \textbf{Train/Val} & \textbf{IW} & \textbf{OW} & \textbf{CR} \\
		\midrule[1pt]
		\multicolumn{5}{l}{\textit{SAMSum}} \\
		{DialIndirect} & 14,731/818 & 124.10 & 157.41 & 1.31 \\
		{ExtSum} & 14,731/818 & 31.23 & 23.44 &0.94  \\
		{ExtSumM} & 14,731/818 & 66.09 &23.44 & 0.69 \\
		{EntSent} & 29,757/1,654 & 31.05 & 11.93 &0.68  \\
		{ExtSentM} & 29,757/1,654 & 46.45 & 11.93 & 0.60 \\
		{DSum} & 14,731/818 & 124.10 & 23.44 & 0.25 \\
		{DialSent} &29,757/1,654  & 149.93 & 11.93 & 0.13 \\
		\midrule[1pt]
		\multicolumn{5}{l}{\textit{DialSumm}} \\
		{DialIndirect} & 12,460/500 & 187.52 & 215.30 & 1.16 \\
		{ExtSum} & 12,460/500 & 44.43 & 30.02 &0.84 \\
		{ExtSumM} & 12,460/500 & 94.32 &31.02 &0.61  \\
		{EntSent} & 22,407/840 & 39.27 &17.78  &0.65  \\
		{ExtSentM} &22,407/840  &61.17  & 17.78 & 0.56 \\
		{DSum} & 12,460/500 & 187.52 & 31.02 & 0.18 \\
		{DialSent} &22,407/840  &214.00 & 17.78 & 0.10 \\
		\bottomrule[1pt]
	\end{tabular}
	\caption{Statistics of constructed datasets. IW and OW refer to the number of words in the input and output of corresponding dataset. DSum and DialSent are in-list for easier comparison.}
	\label{tab:rephrasedatasets}
\end{table}

To make the input and output carry the same amount of information,
one way is to fix $D$ as input and convert utterances into indirect speech
as the output. \citet{ganesh2019restructuring} restructured dialogue into text with complicated rules
%considering discourse relations among utterances for zero-shot scenarios. However, their complicated rules
which are not released and difficult to transfer among datasets under different scenarios. Thus, we only use simple rules to convert all of the utterances into [$r_t$ says,``$u_t$''] and concatenated as the output. We call this dataset as \textbf{DialIndirect}.

Another way is fixing $S$ as output and removing the redundant utterances in $D$ to get the rephrasing input. We take advantage of the idea of oracle extraction for news summarization~\cite{zhou-etal-2018-neural-document} and regard the combination of dialogue utterances with the highest Rouge scores computed with $S$ as the input. 
Considering that utterances are highly dependent, we modify the original extraction algorithm by extracting all of the utterances lying between the extracted ones, 
different from the window-sized snippet selection in~\cite{liu-etal-2021-topic-aware}.  Datasets with or without this modification are called \textbf{ExtSum} and \textbf{ExtSumM} respectively.

A summary $S$ is divided into sentences to construct more rephrase pairs.
%We use Spacy~\footnote{\url{https://spacy.io/}} to
Similar extraction operations can be done between $D$ and $p$, and we 
get \textbf{ExtSent} and \textbf{ExtSentM} datasets.

An example of the paraphrase pair generated from the dialogue-summary pair 
in Figure~\ref{fig:example} is shown
in Table~\ref{tab:datasets2}.
The statistics of post-training datasets derived from SAMSum and DialSumm are shown in Table~\ref{tab:rephrasedatasets}. We compare the performances between different rephrasing approaches with these datasets of our two-stage approach with the fine-tuning-only BART. The results are in Table~\ref{tab:rephrasing}. 

DialIndirect performs incredibly well on SAMSum. However, if we use the converted dialogue as input and directly fine-tune the original BART, the results are only 50.91/28.51/50.25 for Rouge-1/2/L.
It shows that when accompanied with the post-training stage, the model can learn relationships between speakers and utterances, and boundaries of utterances better than a direct transformation of dialogue inputs. 
This rule-based transformation falls on DialSumm compared with BART baseline. More complicated rules may lead to better results, but such labored work is not what we are after. 

\begin{table}[t]
	\scriptsize
	\centering
	\begin{tabular}{lccc}
		\toprule[1pt]
		\textbf{Models} & \textbf{Rouge-1} & \textbf{Rouge-2} & \textbf{Rouge-L} \\
		\midrule[1pt]
		\multicolumn{4}{l}{\textit{SAMSum}} \\
		{BART} &52.06 &27.45 &48.89 \\
		{DialIndirect} &  53.08&28.51  & \textbf{50.25} \\
		{ExtSum} &  53.20& 28.26 & 49.80 \\
		{ExtSumM} & 52.20 & 27.91 & 49.74   \\
		{EntSent} & 51.82 & 27.43 & 49.19   \\
		{ExtSentM} & 51.66 & 27.27 &  48.96  \\
		{DSum} & 52.52&27.51 &49.03 \\
		{DialSent} & \textbf{53.54} & \textbf{28.91} &  50.21 \\
		
		\midrule[1pt]
		
		\multicolumn{4}{l}{\textit{DialSumm}} \\
		{BART} & 53.01&29.18 &51.34 \\
		{DialIndirect} &  52.54&29.13  &51.68 \\
		{ExtSum} & 51.83 & 27.92 &50.33  \\
		{ExtSumM} & 52.29 & 27.72 & 50.09   \\
		{EntSent} & 51.41 & 27.81 & 49.65   \\
		{ExtSentM} & 52.46 & 28.86 & 51.36 \\
		{DSum} & 53.27 & 28.64 & 51.69 \\
		{DialSent} &\textbf{54.73} & \textbf{30.47}&  \textbf{53.46}  \\
		\bottomrule[1pt]
	\end{tabular}
	\caption{Comparisons among different post-training approaches and fine-tuning-only BART baseline on dialogue summarization.}
	\label{tab:rephrasing}
\end{table}

The extraction-based methods fall behind the others. The modification to the algorithm tends to bring more noises than useful information to the input as the results drop mostly. 
Besides, splitting the summary into sentences doesn't improve the results here.
In a word, such hard extractions hurt the intricate discourse and coreference relations among utterances and are not suitable for cross-format data construction.

DialSent with PGG task outperforms other methods and BART consistently across datasets, while DSum with PGG performs almost the same as BART.
If we use DialSent data to augment the original DSum during fine-tuning, the results on SAMSum are 44.61/22.81/44.15 for Rouge-1/2/L respectively showing that the data in both datasets is not compatible. Thus, our approach is different from data augmentation.
Overall, post-training with cross-format rephrasing intuition does help with dialogue summarization,

\section{Case Studies}
\label{sec:cases}
We show more cases as follows.

\begin{table}[th]
	\scriptsize
	\centering
	\begin{tabular}{lp{4.9cm}}
		\toprule[1pt]
		{Dialogue} &  \makecell[l]{\textbf{Kate}: Hey, do you know if our medical \\insurance covers hospital costs? \\\textbf{Greg}: Hm, it depends \\\textbf{Mel}: What happened dear? \\\textbf{Kate}: I broke my arm and they're \\sending me to the hospital :/ \\\textbf{Greg}: Call Linda or ask someone at the \\reception, they should be able to tell \\you what kind of package you have \\\textbf{Kate}: thnx} \\
		
		\hline
		{Reference} & \textbf{Kate} broke her arm and she's going to the hospital. She'd like to know whether her medical insurance covers hospital costs. \textbf{Greg} suggests her to call \textbf{Linda} or ask someone at the reception about it. \\

		\hline
		{BART} &   \textbf{Kate} broke her arm and they're sending her to the hospital. \textbf{Greg} doesn't know if their medical insurance covers hospital costs. \textbf{(53.33/37.93/53.19)}\\
		
		\hline
		{Multi-view} &  \textbf{Kate} broke her arm and they're sending her to the hospital. \textit{\textbf{Greg} will call \textbf{Linda} or ask someone} at the reception to find out if their insurance covers hospital costs.\textbf{(67.64/51.52/56.15)}\\
		
		\hline
		{DialoBART} &   \textbf{Kate} broke her arm and they're sending her to the hospital . \textbf{Greg} advises her to call \textbf{Linda} or ask someone at the reception .\textbf{(65.57/50.85/67.62)}\\
		
		\hline
		{DialSent-PGG} &\textbf{Kate} broke her arm and they're sending her to the hospital. \textbf{Greg} advises her to call \textbf{Linda} or ask someone at the reception if their insurance covers hospital costs. \textbf{(71.64/55.38/62.39)}\\
		\bottomrule[1pt]
	\end{tabular}
	\caption{A case from SAMSum. \textbf{Names} are in bold and \textit{unfaithful contents} are in italic. Rouge-1/2/L scores(\%) are in parentheses.}
	\label{tab:case1}
\end{table}

The case in Table~\ref{tab:case1} is a dialogue happened between three speakers from SAMSum. 
The labeled dialogues, which are directly extracted from Multi-view's and DialoBART's released datasets are shown in Table \ref{tab:case1inputs}. ``$\mid$'' label for Multi-view refers to the topic transitions and stage transitions for the same dialogue respectively. 
We can see that topic segments by Multi-view BART are reasonable. However, such linear segmentation is not quite suitable for this dialogue since the first and third topics are the same.
``$\mid$'' in DialoBART just refers to the end of each utterance.
DialoBART failed to label any topic transitions or redundant utterances.

Compared to the reference summary, the summary generated by BART lost the information about Greg's suggestion, and DialoBART lost the information about ``medical insurance'' even though it recognized ``medical insurance'' as a keyword.
Multi-view did incorrect reasoning on who will call Linda.
Our model generated a more condensed summary covering the same key points as the reference with the original dialogue as input.

\begin{table}[h]
	\scriptsize
	\centering
	\begin{tabular}{lp{5cm}}
		\toprule[1pt]
		\makecell[l]{{Multi-view}\\ {Topic}}& Kate: Hey, do you know if our medical insurance covers hospital costs? Greg: Hm, it depends $\mid$ Mel: What happened dear? Kate: I broke my arm and they're sending me to the hospital :/ $\mid$ Greg: Call Linda or ask someone at the reception, they should be able to tell you what kind of package you have Kate: thnx $\mid$
		\\
		
		\hline
		\makecell[l]{{Multi-view}\\ {Stage}} & $\mid$ Kate: Hey, do you know if our medical insurance covers hospital costs? Greg: Hm, it depends Mel: What happened dear? $\mid$ Kate: I broke my arm and they're sending me to the hospital :/ $\mid$ Greg: Call Linda or ask someone at the reception, they should be able to tell you what kind of package you have Kate: thnx
		\\
		
		\hline
		\makecell[l]{{DialoBART}} & Kate : Hey , do you know if our medical insurance covers hospital costs ? $\mid$ Greg : Hm , it depends $\mid$ Mel : What happened dear ? $\mid$ Kate : I broke my arm and they're sending me to the hospital $\mid$ Greg : Call Linda or ask someone at the reception , they should be able to tell you what kind of package you have $\mid$ Kate : thnx \#KEY\# Mel Kate Greg Hey do you know if our medical insurance covers hospital costs happened dear Linda reception package\\
		
		\bottomrule[1pt]
	\end{tabular}
	\caption{Modified inputs by Multi-view and DialoBART.}
	\label{tab:case1inputs}
\end{table}

Another case from DialSumm between two speakers is in Table~\ref{tab:case3}. BART recognized ``him'' in the second utterance as ``\#Person1\#'' incorrectly. DialoBART regarded the man as ``\#Person1\#'s friends'' which isn't mentioned in the original dialogue.  Our model, DialSent-PGG generates a more accurate summary.

\begin{table}[th!]
	\scriptsize
	\centering
	\begin{tabular}{lp{5cm}}
	\toprule[1pt]
	{Dialogue} & \makecell[l]{\textbf{\#Person1\#}: Like a cat on hot bricks, as \\you might say. I don ' t believe you are\\ listening at all.\\ \textbf{\#Person2\#}: Sorry, I just worried about\\ him. You know, he should be here an \\hour ago. \\ \textbf{\#Person1\#}: Don ' t worry him, he has \\been grown up and I think he can take\\ himself very well. \\ \textbf{\#Person2\#}: But he still does not come\\ back. \\ \textbf{\#Person1\#}: Maybe he is on the way\\ home now.
	} \\
	\hline
	{Reference-1} & \textbf{\#Person2\#} is worried about one man, and \textbf{\#Person1\#} thinks that that man might be on the way home now. \\
	\hline
	{Reference-2} & \textbf{\#Person2\#} is worried about a man, but \textbf{\#Person1\#} thinks it would be fine.\\
	\hline
	{Reference-3} &\textbf{\#Person2\#} is worried about a man but \textbf{\#Person1\#} is not.\\
	\hline
	{BART} & \textit{ \textbf{\#Person2\#} is worried about \textbf{\#Person1\#} }because he hasn't come back from work. \textbf{(43.48/28.57/50.01)} \\
	\hline
	{DialoBART} &  \textbf{\#Person2\#} is worried about \textit{\textbf{\#Person1\#}'s friend }who hasn't come back. \textbf{(45.45/30.00/51.87)} \\
	\hline
	{DialSent-PGG} & \textbf{\#Person2\#} is worried about a boy who hasn't come back.\textbf{(47.62/42.11/53.90)}\\
	\bottomrule[1pt]
	\end{tabular}
	\caption{A case from DialSumm.}
\label{tab:case3}
\end{table}